\documentclass[runningheads]{llncs}

\usepackage{eccv}

\usepackage{eccvabbrv}

\usepackage{graphicx}
\usepackage{booktabs}

\usepackage[accsupp]{axessibility}  

\usepackage{hyperref}

\usepackage{orcidlink}

\usepackage[ruled,vlined]{algorithm2e}

\begin{document}

\title{Explain via Any Concept: Concept Bottleneck Model with Open Vocabulary Concepts} 

\titlerunning{OpenCBM}

\author{Andong Tan\inst{1}\orcidlink{0009-0008-4419-1837} \and
Fengtao Zhou\inst{1}\orcidlink{0000-0001-7039-1156} \and
Hao Chen\inst{1,2,3}\orcidlink{0000-0002-8400-3780}}

\authorrunning{A. Tan et al.}

\institute{Department of Computer Science and Engineering,\\ Hong Kong University of Science and Technology, Hong Kong, China \and
Department of Chemical and Biological Engineering and Center for Aging Science, Hong Kong University of Science and Technology, Hong Kong, China \and
HKUST Shenzhen-Hong Kong Collaborative Innovation Research Institute, Shenzhen, China \\
\email{\{atanac, fzhouaf\}@connect.ust.hk}, \email{jhc@cse.ust.hk}}
\maketitle

\begin{abstract}
The concept bottleneck model (CBM) is an interpretable-by-design framework that makes decisions by first predicting a set of interpretable concepts, and then predicting the class label based on the given concepts. Existing CBMs are trained with a fixed set of concepts (concepts are either annotated by the dataset or queried from language models). However, this closed-world assumption is unrealistic in practice, as users may wonder about the role of any desired concept in decision-making after the model is deployed. Inspired by the large success of recent vision-language pre-trained models such as CLIP in zero-shot classification, we propose ``OpenCBM'' to equip the CBM with open vocabulary concepts via: (1) Aligning the feature space of a trainable image feature extractor with that of a CLIP's image encoder via a prototype based feature alignment; (2) Simultaneously training an image classifier on the downstream dataset; (3) Reconstructing the trained classification head via any set of user-desired textual concepts encoded by CLIP's text encoder. To reveal potentially missing concepts from users, we further propose to iteratively find the closest concept embedding to the residual parameters during the reconstruction until the residual is small enough. To the best of our knowledge, our ``OpenCBM'' is the first CBM with concepts of open vocabularies, providing users the unique benefit such as removing, adding, or replacing any desired concept to explain the model's prediction even after a model is trained. Moreover, our model significantly outperforms the previous state-of-the-art CBM by 9\% in the classification accuracy on the benchmark dataset CUB-200-2011.

\end{abstract}

\section{Introduction}
\label{sec:intro}

Deep models have achieved great success in vision tasks such as image recognition \cite{he2016deep, dosovitskiy2020image, girshick2015fast, ronneberger2015u,bie2024mica, hou2024concept}. However, the high dimensional feature space of deep models is challenging for humans to understand, limiting their real-world deployments. To increase the interpretability of deep models, a rising trend is to design models that are ``interpretable-by-design'', offering humans a transparent reasoning process of the model's prediction \cite{rudin2019stop}. One type of popular model that is ``interpretable-by-design'' is named ``Concept Bottleneck Model (CBM)'', which works by first predicting a set of interpretable concepts and then predicting the class label based on given interpretable concepts \cite{koh2020concept}. These interpretable concepts are represented by high-dimensional features, offering humans a way to interpret the decision-making process of the model in the high-dimensional space. 

\begin{figure}[tb]
  \centering
  \includegraphics[height=6.0cm]{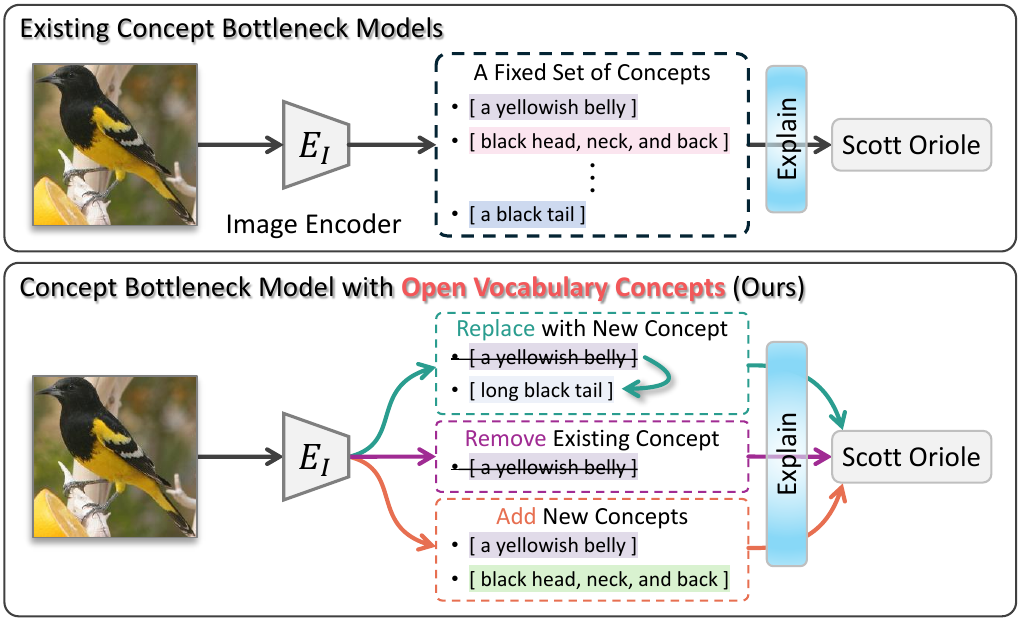}
  \caption{Existing concept bottleneck models are trained with a fixed set of concepts, limiting the users to understand the model's reasoning only according to a static set of concepts. Our OpenCBM offers open vocabulary concepts, allowing users to flexibly choose any set of concepts for the model reasoning without re-training a new model.
  }
  \label{fig:teaser}
\end{figure}

Early CBMs are trained via supervision from the dataset's dense concept annotations, which consumes large human efforts in the dataset preparation \cite{koh2020concept}. More recent CBMs leverage the large language models (LLM) to query a set of task-relevant concepts \cite{oikarinen2023labelfree, yang2023language, yan2023learning,yuksekgonul2023posthoc}, and then predict the class label based on these concepts, extending their applications to scenarios where no concept annotations are available. However, all existing CBMs are trained with a fixed set of interpretable concepts, as described in Figure \ref{fig:teaser}. We note that this closed-world assumption is unrealistic in practice, as users may wonder about the role of any desired concept even after the model is trained and deployed. For example, assume a model is trained to classify an image into different bird sub-species according to a set of concepts such as features of wings and heads, a user may wonder (1) after removing some concepts, whether the remaining concepts could have different importance contributions to the downstream task to dynamically study the concept correlations; (2) whether it is possible to classify the image according to additional concepts (e.g., characteristics in the belly); (3) how would the model behave if replacing the concepts relevant to wings and/or heads with another set of concepts to compare which set of concepts is better? All the above practical scenarios are impossible using existing CBMs.

To address this concern, we propose the ``OpenCBM'' model, which is the first CBM offering open vocabulary concepts. After our ``OpenCBM'' is trained, users can query the ``OpenCBM'' to predict any set of desired concepts, and then classify the images based on this set of concepts. This offers the users the complete freedom to remove, add or even completely replace a set of concepts without re-training the model. This benefit has a huge potential impact as users could leverage the open vocabulary concepts for a wide range of applications (e.g., knowledge discovery, model debugging/debias). Inspired by the large success of large-scale pre-trained vision-language models such as CLIP \cite{radford2021learning}, which offers an aligned feature space for images and texts, we propose to align the feature space of a trainable feature extractor to that of the CLIP's image encoder. The alignment aims to enable the human interpretable concept input in the form of texts to be meaningful in the visual feature space. Simultaneously, we train an image classifier on the downstream dataset. Then we reconstruct the trained classifier via any user-desired concept set encoded by the CLIP's text encoder. These concepts can detect relevant features in the images thanks to the shared space between CLIP's image and text encoder. Furthermore, to reveal potentially missing concepts from the user-specified set to recover the model's performance, we further propose an iterative mechanism to find the nearest concept embedding to the residual parameters during the reconstruction until the residual is small enough. Our contributions are summarized as follows:
\begin{itemize}
    \item We propose ``OpenCBM'', the first Concept Bottleneck Model with open vocabulary concepts, allowing the model to use any set of concepts specified by the users for the final prediction even after the model is trained.
    \item We further propose to iteratively find the nearest concept embedding from an interpretable search space to reveal the concepts missed by the user to recover the model's performance.
    \item We demonstrate that our model not only offers unique interpretability advantages to the user but also outperforms the state-of-the-art CBM models by 9\% in the accuracy on the benchmark dataset.
\end{itemize}

\section{Related Work}

\paragraph{Concept Bottleneck Models.}
The CBM \cite{koh2020concept} is an interpretable-by-design framework that makes decisions by first predicting a set of interpretable concepts, and then predicting the class label based on the given concepts. This type of model has been shown to be beneficial by allowing users to conduct test-time intervention or easier debugging \cite{koh2020concept}. To get rid of concept annotations for each input image required to train a CBM, Post-hoc CBM \cite{yuksekgonul2023posthoc} proposes to train a concept bank in a different dataset in advance to obtain concept activation vectors, and then project training data's features to this concept space to construct the CBM. Label-free CBM takes one step further and completely removes the necessity of concept annotations by querying LLM for task-relevant concepts and using a pre-trained vision-language model (e.g., CLIP) to calculate the image-concept similarity to guide the learning of CBM \cite{oikarinen2022label}. LaBO constructs a CBM directly leveraging frozen CLIP's image and text encoder by only learning a class-concept weight matrix \cite{yang2023language}. Since this method requires a huge number of concepts (e.g., 10000 concepts for a dataset of 200 classes), some later work attempts to find a smaller set of more precise and descriptive concepts \cite{yan2023learning}. However, the concept set of all the above works is fixed after the model is trained, making it impossible for users to freely modify the concept set without re-training a new model. Moreover, all the above models under-perform their black-box counterparts in the classification accuracy.

\paragraph{Post-hoc explanations}
This type of work aims to design methods for explaining the model after the model has already been trained. They have the advantage that the model's training process will not be influenced and therefore the model being explained will maintain its accuracy. Typical methods include perturbations based methods \cite{ribeiro2016should, zeiler2014visualizing, zhou2015predicting} and back propagation based methods \cite{selvaraju2017grad, zhang2018top,sundararajan2017axiomatic}. However, unlike Concept Bottleneck Models, these methods have no guarantee to faithfully represent the model's reasoning process \cite{rudin2019stop}.

\section{Method}

\subsection{Preliminaries on Concept Bottleneck Models \label{sec:preliminary}}
Denote an input image as $x$, a concept set as $z$, and the class label as $y$, the concept bottleneck model enforces the model to first predict interpretable concepts $z$ based on the input $x$ via a network $f$ and then predict the class label $y$ using only the interpretable concepts via another network $h$. The overall prediction chain can be expressed as $x\rightarrow z \rightarrow y$, where $y=h(f(x))$. A standard image classification model can also be expressed in a similar form if defining $f$ to be the feature extractor and $h$ as a classifier. We point out that standard models will be equivalent to a concept bottleneck model if the classifier $h$ could be decomposed into two parts $h=h_1 \circ h_2$, where $h_1$ predicts interpretable concepts and $h_2$ predicts the class label based on concepts. The notion of $\circ$ means an input sequentially passing two networks $h_1$ and $h_2$ will be equivalent to passing a single network $h$. After the decomposition, $f\circ h_1$ will be the concept predictor and $h_2$ will be the network predicting class label. In this work, we propose to decompose the final linear classifier via a set of text embedding that represent any user-desired concepts. The similarities of input features to these concepts serve as the concept predictor and the coefficients of these concepts indicate a network with a single fully connected layer predicting the class label using predicted concepts. This simple form of linear combinations has the interpretability advantage that the contribution of each individual concept can be clearly identified, as commonly applied in prior works \cite{alvarez2018towards,oikarinen2022label, yang2023language}.

\subsection{Prototype Based Feature Space Alignment}

\begin{figure}[tb]
  \centering
  \includegraphics[height=7.5cm]{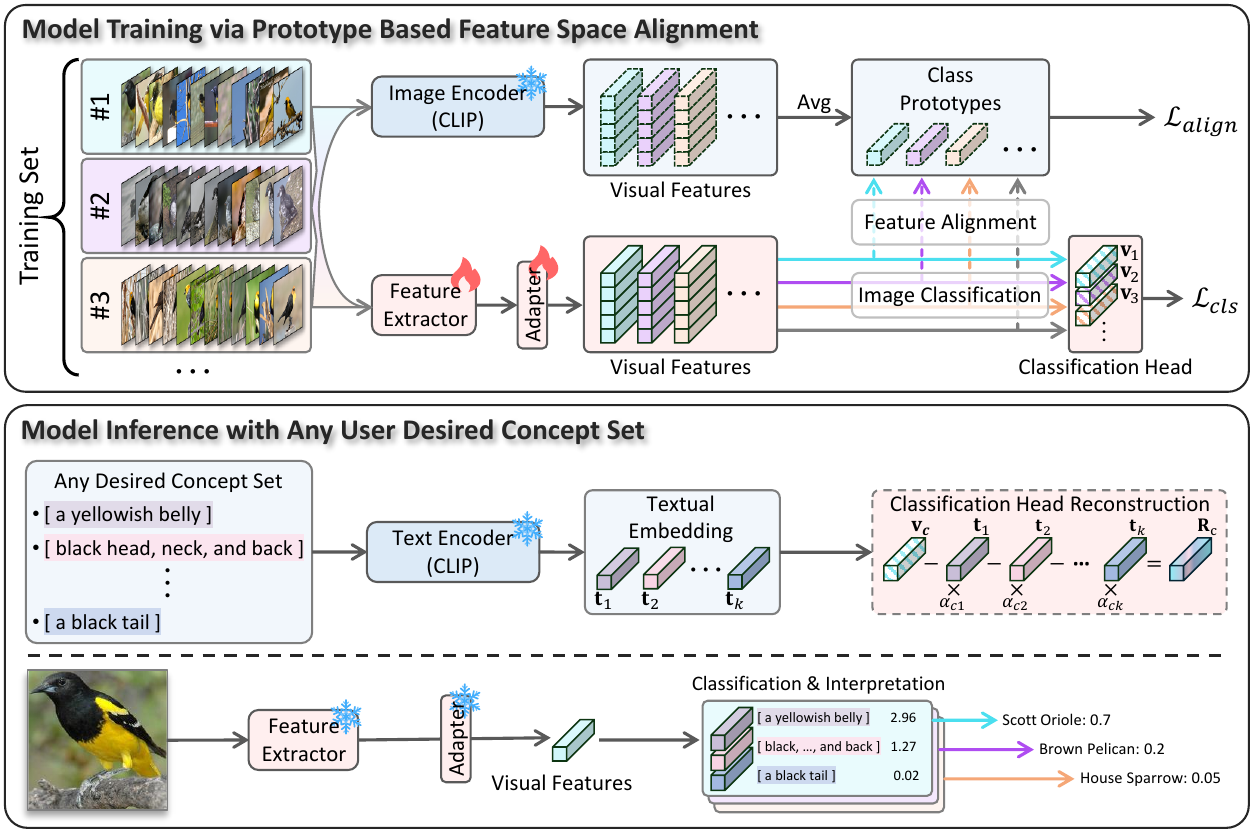}
  \caption{Our method works by training a standard network (e.g., ResNet) while aligning its feature space with that of the CLIP's feature space via a class prototype based feature space alignment. At inference time, we propose to leverage any user desired concept set (e.g., $k$ cocepts ) encoded by CLIP to reconstruct the trained classification head $\mathbf{v}_c$. The residual parameters after the reconstruction encode the missing concepts not queried by the user.
  }
  \label{fig:framework}
\end{figure}

We denote $C$ as the number of classes, $D$ as the feature dimension, $i$ as the index of an image, $\mathbf{x}_i \in \mathbb{R}^{HW\times 3}$ as the image input with height $H$ and width $W$, $N_c$ as the number of images belonging to the class $c$ and $k$ is the number of concepts desired by the user. To leverage the shared feature space of vision and text encoder offered by vision-language pre-trained (VLP) models such as CLIP \cite{radford2021learning}, we propose to align the feature space of a trainable feature extractor with that of the CLIP's image encoder, such that the CLIP's text encoder also shares with the same feature space as our trainable feature extractor. This enables the human interpretable concepts expressed in texts/languages to have corresponding visual meaning in the image space to construct the concept bottleneck model. We incorporate a single linear layer as the adapter to match the feature dimension of any feature extractor to the dimension of any pre-trained vision-language model. We train an additional feature extractor instead of directly using the one offered by CLIP to more flexibly fit to the specific downstream task. Therefore, the trainable feature extractor serves to map training data to the CLIP's feature space, yet makes the extracted features more class-discriminative to the downstream classification task. 

To achieve this goal, we propose a prototype based feature space alignment, which aligns each image towards the prototype of its corresponding class, as shown in Figure \ref{fig:framework}. The prototype is calculated by feeding all training images of the same class to the CLIP's image encoder and simply averaging their features. We do not directly align the trainable feature extractor's feature with the one extracted by a frozen CLIP for each input sample to avoid aligning with features not well extracted by CLIP (e.g., some features of one class may reside in the areas dominated by features of another class). In this way, we obtain a set of prototypes equal to the number of classes, which describe the feature space that the downstream task's visual features should roughly live in. The prototype $\mathbf{P}_c \in \mathbb{R}^{D}$ for each class $c$ is calculated as follows:
\begin{equation}
    \mathbf{P}_c= \sum_i^{i=N_c} f_{clip}(\mathbf{x}_i)/N_c, \forall \mathbf{x}_i \in  \mathcal{X}_c,
\end{equation}
where $f_{clip}$ indicates the frozen CLIP's image encoder and $\mathcal{X}_c$ denotes all images belonging to the class $c$. The alignment is encouraged via a cosine loss function:
\begin{equation}
\label{eq:alignment}
 \mathcal{L}_{align} =  -
 \frac{ \mathbf{P}_c(g(\mathbf{x}_i))^T}{\|\mathbf{P}_c\|\|g(\mathbf{x}_i)\|}, \forall \mathbf{x}_i \in \mathcal{X}_c,
\end{equation}
where $g$ denotes the trainable feature extractor. We use cosine similarity as the CLIP is also trained via cosine similarity to align the feature space between the vision and the text.

In parallel to the feature space alignment, the classification head is simultaneously trained via a standard cross entropy loss \cite{shannon1948mathematical}, denoted by $ \mathcal{L}_{cls}$. Therefore, overall the training loss $\mathcal{L}$ is expressed as:
\begin{equation}
\mathcal{L} = \beta_1\mathcal{L}_{cls} + \beta_2 \mathcal{L}_{align}.
\end{equation}
$\beta_1$ and $\beta_2$ are hyperparameters to adjust the contributions of two loss functions.

\subsection{Obtaining the Importance of Any Desired Concept\label{sec:concept_importance}}
This is represented by the optimized coefficients after approximating the classification head of each class via the set of desired concepts' embedding linearly. The approximation is meaningful as the CLIP's text encoder and our trained classifier shares the same feature space after the alignment. The optimization objective is:
\begin{equation}
\label{eqa:cls_err}
    \min_{\alpha_{ci}} \sum_{c=1}^{C}||\mathbf{v}_c - \sum_{i=1}^{k}\alpha_{ci}\mathbf{t}_i||_2^2,
\end{equation}
where $\alpha_{ci}$ indicates the importance of $i^{th}$ concept $\mathbf{t}_i \in \mathbb{R}^D$ for class $c$ and $\mathbf{v}_c\in \mathbb{R}^D$ denotes the classification head corresponding to the class $c$. $\|\|$ indicates the Frobenius norm. To this end, the user could immediately know the importance of any desired concept set without training a new neural network.

\subsection{Discovery of Missing Concepts \label{sec:interpret_residual}}
When the user offered concepts are not class-discriminative enough, the model's performance might be low if enforcing the model to only rely on these concepts (e.g., in the extreme case, a user may input a set of task-irrelevant concepts). In this case, we further propose an iterative mechanism to reveal the missing concepts $\mathbf{T}_{miss}$ to fully recover the performance of our trained ``OpenCBM''. After the coefficients $\alpha_{ci}$ are optimized as introduced in section \ref{sec:concept_importance}, the residual parameters $\mathbf{R}_c \in \mathbb{R}^{1\times D}$ in each class $c$ are calculated as:
\begin{equation}
    \mathbf{R}_c = \mathbf{v}_c - \sum_{i=1}^k \alpha_{ci} \mathbf{t}_i.
\end{equation}
These parameters encode the information delivered by missing concepts. If $\mathbf{R}_c$ is small enough, the existing concept set already well covers necessary concepts. If not, to further interpret these parameters, we propose to iteratively find the nearest concept embedding from a concept search space $\mathbf{T}$ to these residual parameters until the difference is smaller than a threshold $\epsilon$, as described in Alg. \ref{alg:discovery}. The search space could be any large concept set specified by the user. This can for example be created via some known knowledge graph such as ConceptNet \cite{speer2017conceptnet}, or by simply asking large language models to generate a sufficiently large concept search space. We note that there exists an unlimited number of possible linear combinations of high-dimensional vectors that could reconstruct the residual parameters, and most of them are even non-interpretable. However, as pointed out by \cite{rudin2022interpretable}, interpretable machine learning is exactly aiming to find one model that is interpretable among all possible models (this huge model space is also named as ``Rashamon set'') that could reach the same accuracy on a given dataset. In our context, this means finding one set of vectors corresponding to interpretable concepts' embedding to reconstruct the classification head.

\begin{algorithm}[H]
\label{alg:discovery}
\SetAlgoLined
\KwIn{Residual parameters $\mathbf{R}_c$, concept search space $\mathbf{T}$, tolerance $\epsilon$}
\KwOut{Missing concept set $\mathbf{T}_{miss}$}
Initialization: $\mathbf{R}= \mathbf{R}_c$, $\mathbf{T}_{miss}=\emptyset$\;

\While{$\|\mathbf{R}_c\|_2^2>\epsilon$  and $\mathbf{T}$ is not empty}{
    Compute the absolute of cosine similarities $ | \frac{\mathbf{R}_c\mathbf{T}^T}{\|\mathbf{R}_c\| \| \mathbf{T}\|^T} |$ \;
    Find the concept $\mathbf{t}_m \in \mathbf{T}$ with largest value in the last step \;
    Find optimal scaler $\Tilde{\alpha}_{cm}$ for the optimization: $ \min \|\mathbf{R}_c - \alpha_{cm} \mathbf{t}_m \|_2^2$\;
    Update residuals $\mathbf{R}_c \leftarrow \mathbf{R}_c - \Tilde{\alpha}_{cm} \mathbf{t}_m$ and remove $\mathbf{t}_m$ from $\mathbf{T}$\;
    Add $\mathbf{t}_m$ to the missing concept set $\mathbf{T}_{miss}$\;
}
\caption{Discovery of missing concepts.}
\label{alg:gradient_descent}
\end{algorithm}

\subsection{Concept Removal from an Unknown Concept Set\label{sec:remove_from_unknown}}

In typical scenarios, concept removal indicates removing the influence of a certain concept from a given concept set. This could be easily achieved by simply changing the concept set and applying the optimization introduced in section \ref{sec:concept_importance}. In this part, we show a different scenario: our ``OpenCBM'' has the unique advantage of allowing the user to directly remove a desired concept from an unknown concept set. Conventional test-time user intervention in existing CBMs can not achieve this because users can only intervene in concepts that the model is explicitly trained to predict. 

Concretely, we propose to directly remove the feature dimension corresponding to the user's desired concept from the classification head of each class. The classification head vector in each class encodes a complete set of concepts to reach the corresponding accuracy. Therefore removing one dimension specified by the concept's text encoding allows the model to not rely on the removed concept. This enables the user to directly remove a concept without needing to specify a concept set in advance. An example usage is: in bird species classification, existing CBMs are never trained to predict concepts such as the sky or trees, making it impossible to intervene in these concepts. However, our ``OpenCBM'' can easily remove these concepts and identify to which extent the model's reasoning relies on these background concepts by observing the accuracy change using the modified classification head. To fully remove the dimension along a desired concept $\mathbf{t}_d$ axis for the reasoning of each class $c$, we propose to remove the concept weighted by a scale $\gamma_c \in \mathbb{R}$, which is obtained via:
\begin{equation}
\label{eqa:remove}
    \min_{\gamma_c} \sum_{c=1}^C ||\mathbf{v}_c - \gamma_c\mathbf{t}_d||_2^2.
\end{equation}
After the above optimization, the new classification head $\Tilde{\mathbf{v}}_c \in \mathbb{R^D}$ for each class $c$ becomes:
\begin{equation}
    \Tilde{\mathbf{v}}_c = \mathbf{v}_c-\Tilde{\gamma}_c\mathbf{t}_d,
\end{equation}
where $\Tilde{\gamma_i}$ is the optimized scale parameter. The model's behavior after the concept removal could be observed via applying the newly obtained classification head consisting of $\Tilde{\mathbf{v}}_1, ..., \Tilde{\mathbf{v}}_C$.

\subsection{Concept Adding, Removal and Replacement in a Known Concept Set}
Given a known concept set, our ``OpenCBM'' can flexibly add, remove, and replace concepts. The importance of concepts in the modified set of concepts can be easily re-calculated by conducting the reconstruction introduced in the section \ref{sec:concept_importance}. If the reconstruction is perfect (e.g., with negligible approximation error), a CBM is well constructed with the given concepts. If the reconstruction is not accurate enough, one could further leverage the concept discovery mechanism proposed in section \ref{sec:interpret_residual} to discover more concepts when the user wants the decision making be grounded only on known concepts. The user can also simply leave the residual parameters as they are as a representation encoding all missing concepts if being only interested in the existing concepts.

\subsection{Overview of the Inference Process}
Given the extracted feature $g(\mathbf{x}_i)$, when only using the known queried concept set, the class logit of class $c$ can be calculated as:
\begin{equation}
     \alpha_{c1}g(\mathbf{x}_i)\mathbf{t}_1^T + ... +\alpha_{cj}g(\mathbf{x}_i)\mathbf{t}_j^T+ ...+\alpha_{ck}g(\mathbf{x}_i)\mathbf{t}_k^T.
\end{equation}
The coefficients $\alpha_{cj}$ indicates the importance of the $j^{th}$ concept regarding the class $c$. These coefficients could be understood as the parameters of the class label prediction network $h_2$ and all the $g(\mathbf{x}_i)\mathbf{t}_j^T$ can be understood as a concept prediction network $h_1$ as introduced in section \ref{sec:preliminary}. The full performance of our model could be obtained via:
\begin{equation}
g(\mathbf{x}_i)\mathbf{v}_c =        \alpha_{c1}g(\mathbf{x}_i)\mathbf{t}_1^T + ... +\alpha_{cj}g(\mathbf{x}_i)\mathbf{t}_j^T+ ...+\alpha_{ck}g(\mathbf{x}_i)\mathbf{t}_k^T +g(\mathbf{x}_i) \mathbf{R}_c,  
\end{equation}
where $\mathbf{R}_c$ encodes the concepts not explicitly queried by the users.

\section{Experiments}
In this section, we first demonstrate the performance advantage of our method compared to prior works. Then we show a case study of flexible concept editing, which is a unique capability of our ``OpenCBM'' model. A comprehensive ablation study is offered at the end. 

\subsection{Classification Accuracy}

\textit{Implementation details.} We leverage the ImageNet \cite{deng2009imagenet} pre-trained models as initialization for the backbone following \cite{koh2020concept}, and replace the classification head with a randomly initialized linear classifier capable of classifying 200 classes to fit the dataset class number. We train 100 epochs with an initial learning rate of 0.001 and multiply it by 0.1 every 30 epochs. We use SGD as the optimizer with momentum 0.9 and weight decay $10^{-4}$. We only use horizontal flip with probability 0.5 as the data augmentation. We set $\beta_1 =1$ and $\beta_2=5$ unless otherwise indicated. 

We compare our ``OpenCBM'' with existing state-of-the-art CBMs in the classification accuracy on the benchmark dataset CUB-200-2011 \cite{wah2011caltech}. This is a dataset for bird species classification (200 classes). We follow the default training and validation set split for the evaluation. We use the ResNet18 \cite{he2016deep} as the backbone and CLIP-RN50 \cite{radford2021learning} for a fair comparison to prior works. As shown in Table \ref{tab:cbm_acc}, our ``OpenCBM'' significantly outperforms the previous state-of-the-art CBM by 9\%. Even compared to the standard black box ResNet18 \cite{he2016deep} (we also incorporate an adapter before the final layer for fair comparisons), our model has a significant performance boost of 3.4\%. This result indicates that our proposed prototype based feature space alignment not only equips our model with the capability for open vocabulary concepts, but also enables CLIP's pre-trained prototype features to guide the learning of the trainable feature extractor towards better accuracy. We refer to the appendix for more results in CIFAR100 \cite{krizhevsky2009learning}.

To further validate our motivation that a trainable feature extractor may fit better to the downstream task compared to a frozen CLIP's image encoder, we compare our model with the performance of a frozen CLIP-RN50 \cite{radford2021learning} with linear probing. Linear probing is a common technique to evaluate the quality of the pre-trained image encoder towards the downstream task \cite{radford2021learning}. The linear probing is implemented by scikit-learn's L-BFGS logistic regression following \cite{yang2023language}. The boosted accuracy result in Table \ref{tab:cbm_acc} (83.3 \% vs. 75.5\%) validates our design choice. 

\begin{table}[tb]
  \caption{Classification accuracy comparisons of different methods on the benchmark dataset CUB-200-2011 \cite{wah2011caltech} using the ResNet18 \cite{he2016deep} backbone. 
  }
  \label{tab:cbm_acc}
  \centering
  \begin{tabular}{@{}cccc@{}}
    \toprule
    Methods &  Concept annotation & Open vocabulary &Accuracy(\%) \\
    \midrule
    Black box Res18 \cite{he2016deep} & - & - & {79.9}\\
    CLIP-RN50 with linear probing \cite{radford2021learning} & - & - & 75.5 \\
    \midrule
    CBM \cite{koh2020concept} & $\surd$ & $\times$ & {62.9}\\
    P-CBM \cite{yuksekgonul2022post}  & $\surd$ & $\times$& {58.8}\\
    P-CBM-h \cite{yuksekgonul2022post} & $\surd$ & $\times$& {61.0}\\
    LaBO \cite{yang2023language} & $\times$ & $\times$ & {71.8}\\
    Label-free CBM \cite{oikarinen2022label} & $\times$ &$\times$ & {74.3}\\
    OpenCBM (ours)  & $\times$ & $\surd$ & \bf{83.3}\\
  \bottomrule
  \end{tabular}
\end{table}

\subsection{Case Study on the Concept Set Modification}
In this section, we demonstrate the benefits of our method in the open vocabulary setting, where users are completely free to use any desired concept set to explain the model's prediction and interact with the model.

\begin{figure}[tb]
  \centering
  \includegraphics[height=4.5cm]{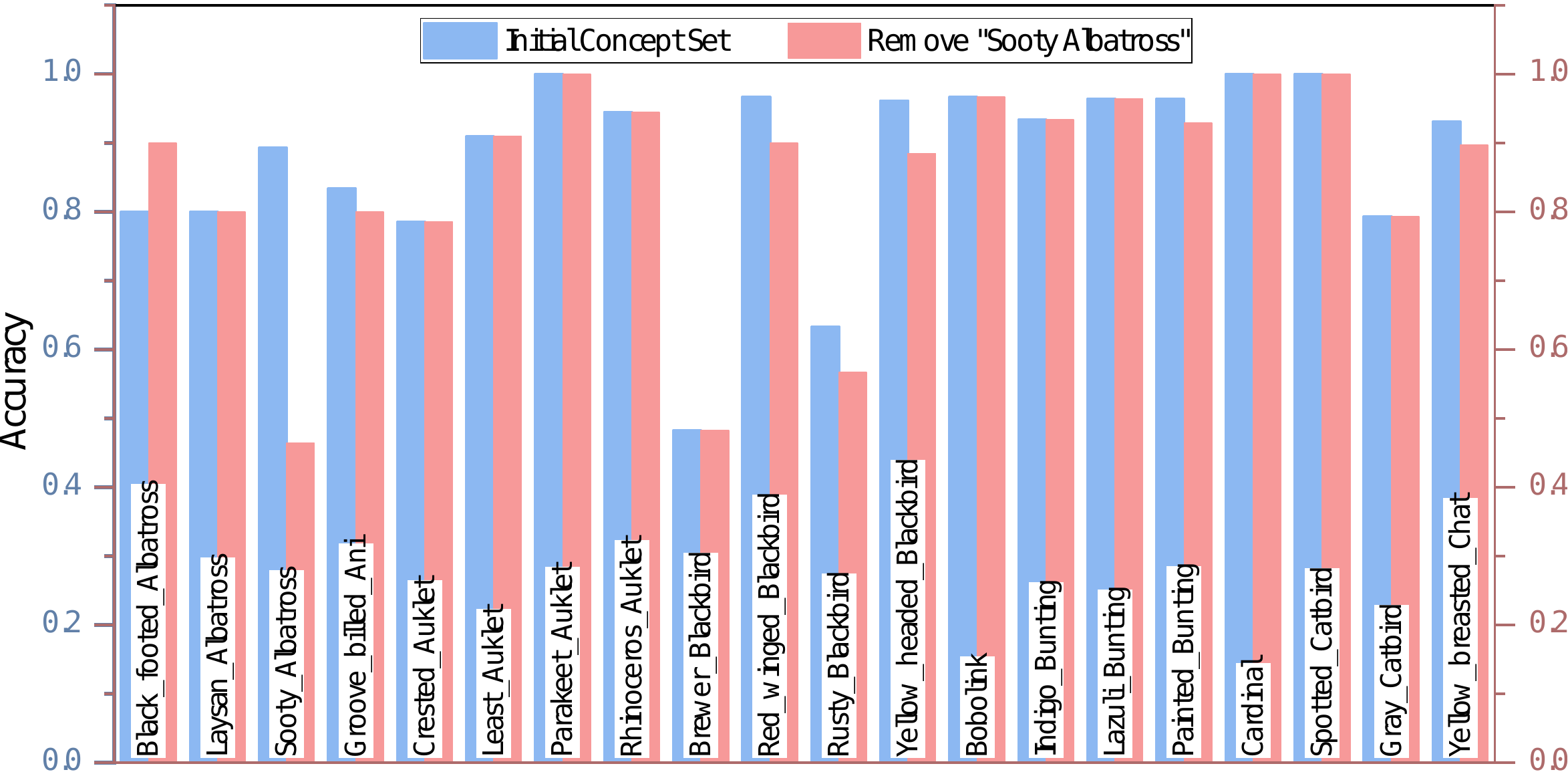}
  \caption{Accuracy change of 20 classes after removing the class name ``Sooty Albatross'' using the technique of section \ref{sec:remove_from_unknown}. The third pair of bars shows a large accuracy drop in ``Sooty Albatross''. The changes in other classes demonstrate the inference correlation between classes, which are positively or negatively correlated with the given class.
  }
  \label{fig:class_wise_change}
\end{figure}

\textit{How well is the feature space of the trainable feature extractor aligned with CLIP's text embedding space?} We answer this question by leveraging the model's unique capability to remove concepts from an unknown concept set. Since the class name is typically observed to be class-discriminative when used alone in zero-shot image classification \cite{saha2024improved}, we set the class name as the desired concept to be removed from an unknown concept set as described in section \ref{sec:remove_from_unknown} and expect a performance drop in corresponding classes. We conduct the removal 200 times for the removal of 200 class names and report the number of classes that exhibit a performance drop when their own class name is removed. We report the average number with standard deviation over 3 checkpoints of our models: $173.6\pm 4.3$ out of 200 classes exhibits a performance drop, validating the effect of our feature space alignment. An illustrative figure demonstrating the accuracy change in different classes when removing the class name ``Sooty Albatross'' is offered in Figure \ref{fig:class_wise_change}. This result reveals the correlations between classes, as the inference of some classes may be helpful for the inference of other classes. Note that not all classes have a performance drop when their own class name is removed as a concept. This is due to the fact that even in the CLIP's feature space, the class name alone is not discriminative enough for a high zero-shot accuracy in the challenging fine-grained image classification task of CUB-200-2011 \cite{wah2011caltech}, where the zero-shot classification accuracy is only 50.54\% even when using a larger version of CLIP (e.g., ViT-B/32) \cite{saha2024improved}. We refer to the appendix for the evaluation of more datasets.

\textit{Adding concepts increases the prediction accuracy.}
Our ``OpenCBM'' offers the flexibility to allow users to add more concepts for the explanation by simply using more concepts' text embedding to approximate the trained classification head. We simulate this scenario by randomly sampling an increasing percentage of concepts from the concept set discovered by LaBO \cite{yang2023language}, which is generated by LLM. We compare the classification accuracy as well as the reconstruction error calculated via equation \ref{eqa:cls_err} in Table \ref{tab:add}. The table shows that using more concepts helps to reduce the approximation error and leads to better accuracy. Note that a smaller approximation error indicates there is less important information missed by the user. The accuracy results also reveal that there does exist redundancy in the concept set, as using 20 \% of the concepts readily yields an accuracy significantly higher than original works (82.94\% vs. 71.8\% ). This finding is consistent with \cite{yan2023learning}, which makes efforts to reduce the size of the concept set.

\begin{figure}[tb]
  \centering
  \includegraphics[height=5cm]{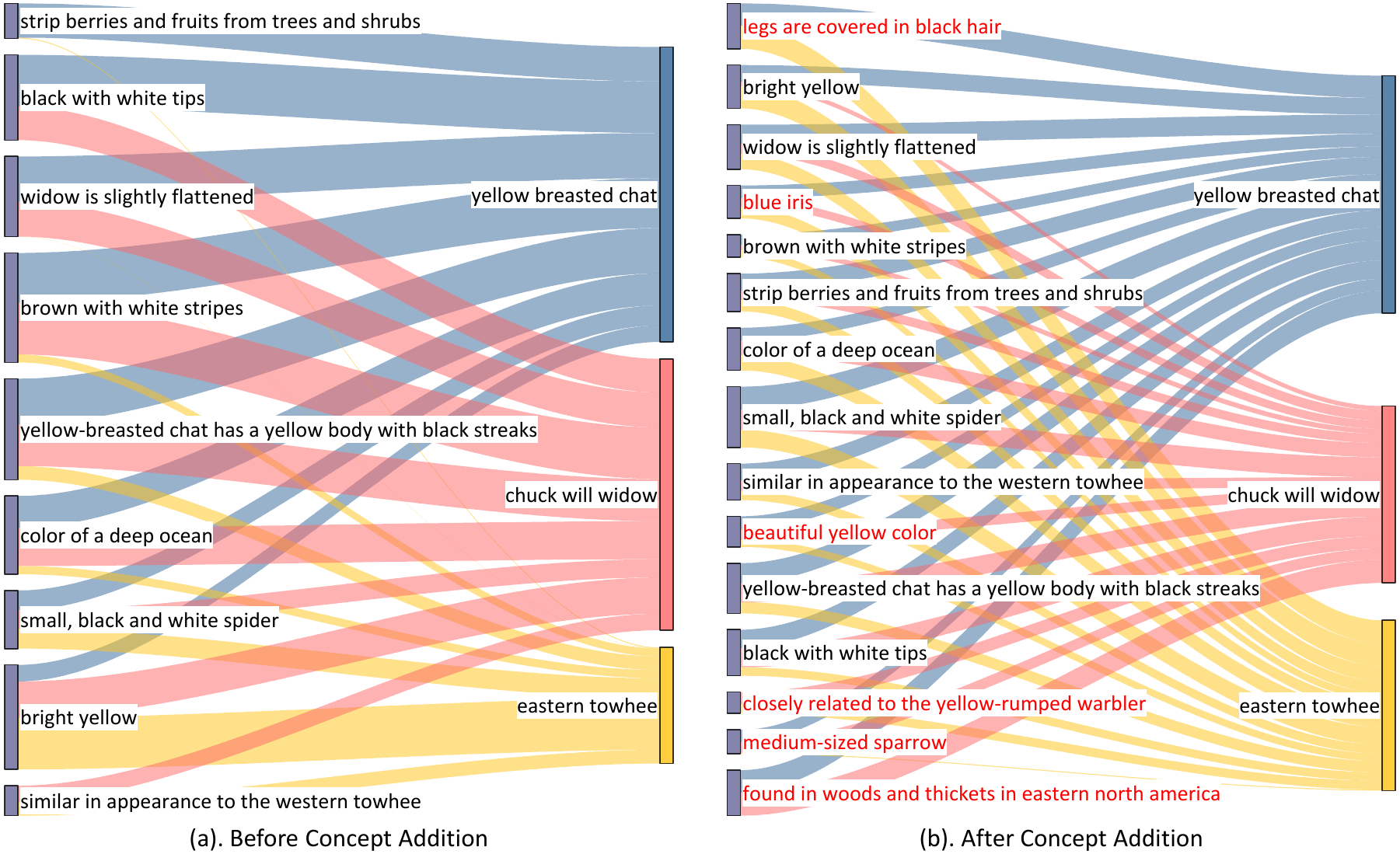}
  \caption{Visualizations of the learned concept importance for different classes. Original concept set has on average 3 concepts per class. After the adding, each class has 5 concepts. Concepts in red colors are newly added.
  }
  \label{fig:concetp_class_relation}
\end{figure}

\begin{figure}[tb]
  \centering
  \includegraphics[height=5.5cm]{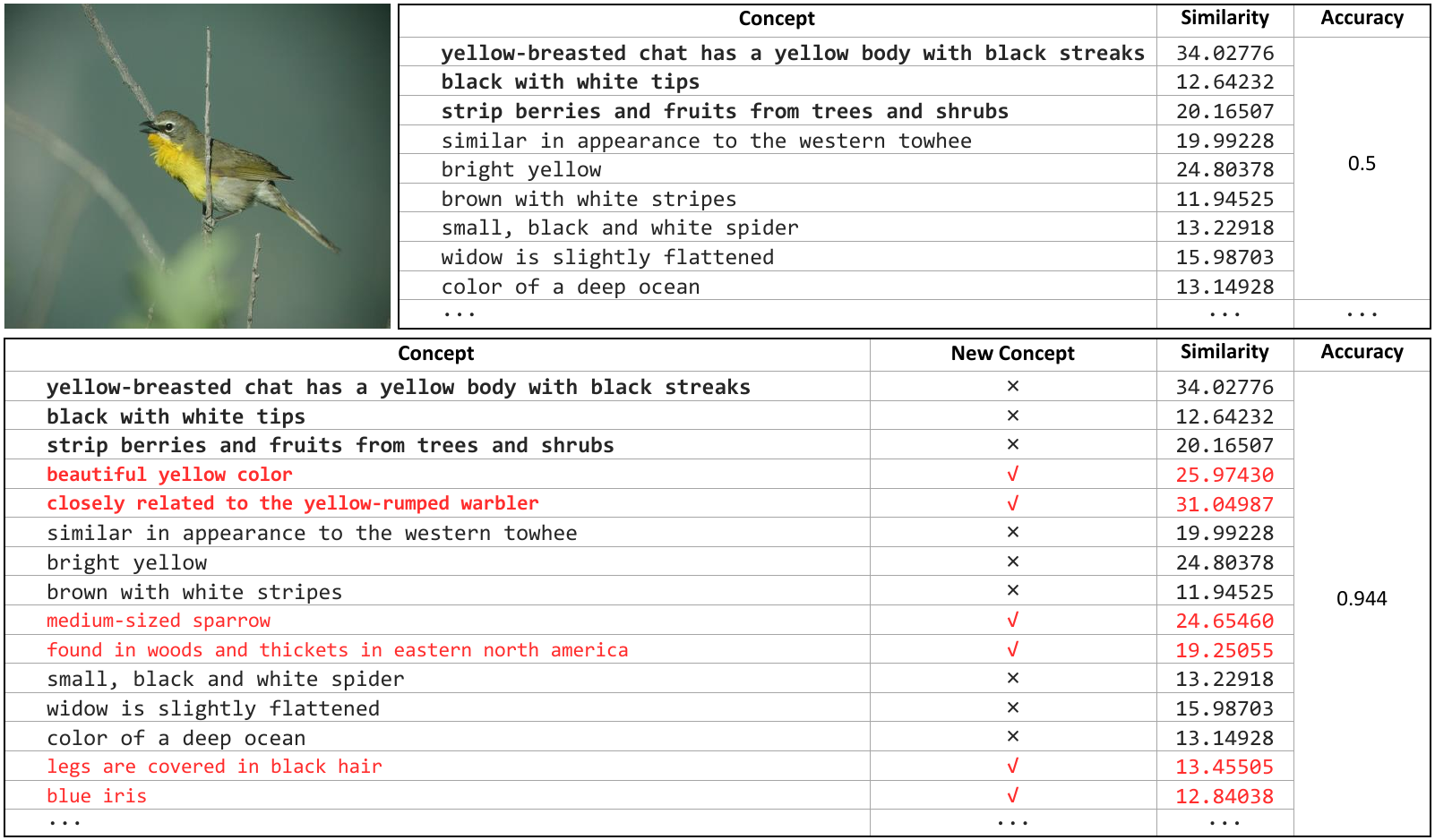}
  \caption{Illustration of an inference process after adding more concepts. Concepts in bold are generated via asking LLM to generate relevant concepts to the class ``yellow-breasted chat''. Red colors indicate newly added concepts. 
  }
  \label{fig:case_study_similarity_table}
\end{figure}

\textit{In-depth analysis of concept addition.} In LaBO's explored concept set \cite{yang2023language}, 50 concepts are discovered for each class. We use the first 3 concepts for each class as the original concept set and compare the difference when using the first 5 concepts for each class to further evaluate the effect of concept addition. Figure \ref{fig:concetp_class_relation} illustrates how adding new concepts will change the coefficients of existing concepts and what are the coefficients of new concepts. To further demonstrate how the model infers with a larger concept set regarding a specific input (e.g., an image of yellow breasted chat), we show the raw visual feature similarity to individual concepts in Figure \ref{fig:case_study_similarity_table}. The result shows that although all concepts together contribute to a better prediction accuracy, the concepts (e.g., yellow-breasted chat has a yellow body with black streaks) originally generated for the corresponding class (concept names in bold) still play most important roles in this prediction. After adding new concepts, the accuracy of the class ``yellow breasted chat'' increases from 50\% to  94.4\%.

\begin{table}[tb]
  \caption{Classification accuracy comparisons after adding concepts. We simulate this scenario via incrementally increasing the percentage of randomly sampled concepts from the concept set explored by the LaBO \cite{yang2023language}. 
  }
  \label{tab:add}
  \centering
  \begin{tabular}{@{}c@{\hspace{0.5cm}}c@{\hspace{0.5cm}}c@{\hspace{0.5cm}}c@{\hspace{0.5cm}}c@{\hspace{0.5cm}}c@{}}
    \toprule
    &5\% & 6\%  & 8\% & 10\% &  20\%  \\
    \midrule
    Accuracy & 44.1  &61.1 & 64.0 &80.3 & 82.94 \\
    Appro. Error & 287.67 & 232.67 & 138.1 &68.1 &32.35 \\
  \bottomrule
  \end{tabular}
\end{table}

\subsection{Missing Concepts Discovery}
If the concept set defined by the user is insufficient, we may automatically discover the missing concepts to fully recover the performance. To achieve this, we leverage the mechanism proposed in section \ref{sec:interpret_residual}. The concept set defined in Label-free CBM \cite{oikarinen2023labelfree} is used to simulate the initial user-queried concepts. We leverage the concept set of LaBO \cite{yang2023language} as one search space and generate another concept set consisting of 2000 concepts ourselves by querying LLM to list 10 most important features for each class as the second search space. Take the class ``yellow breasted chat'' as an example, we interestingly find that the first several concepts revealed in both search spaces include concepts relevant to the habitat or non-visual class features (e.g.,``tree'' from the LaBO concept set and ``melodious song'' from our generated concept set). This indicates that the model is trying to close the approximation gap via more diverse concepts when given visual features' embedding is not accurate enough during the approximation. This capability is potentially helpful to inspire experts for knowledge discovery.

\subsection{Ablation Study}
In this section, we validate our design choices such as prototype based feature space alignment, analyze the prototype correlations, extracted feature distributions, as well as the influence of the alignment strength.

\begin{figure}[tb]
  \centering
  \begin{subfigure}{0.55\linewidth}
    \includegraphics[height=3.5cm]{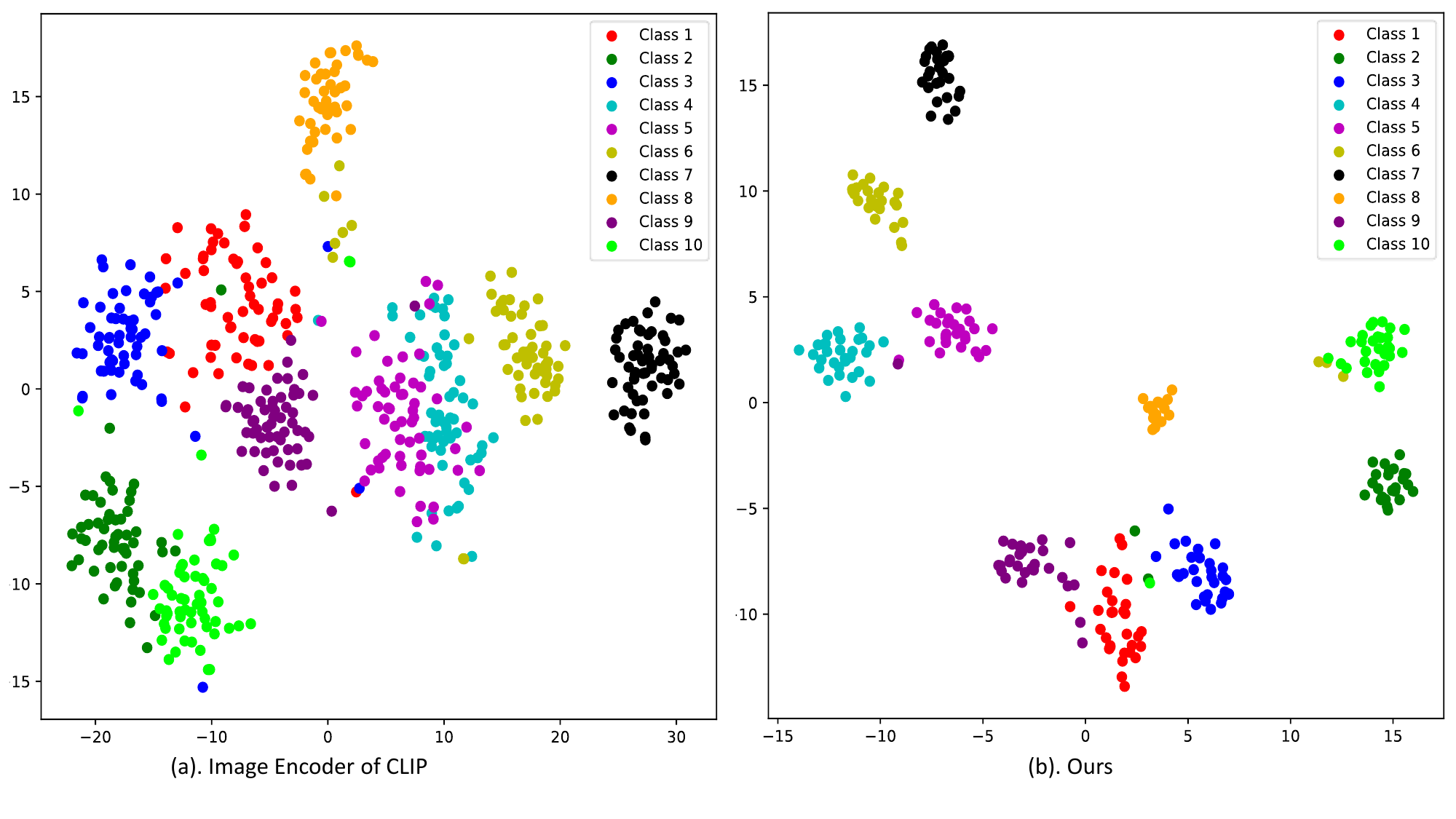}
    \caption{t-SNE visualization of extracted features.}
    \label{fig:t_sne}
  \end{subfigure}
  \hfill
  \begin{subfigure}{0.33\linewidth}
    \includegraphics[height=3.2cm]{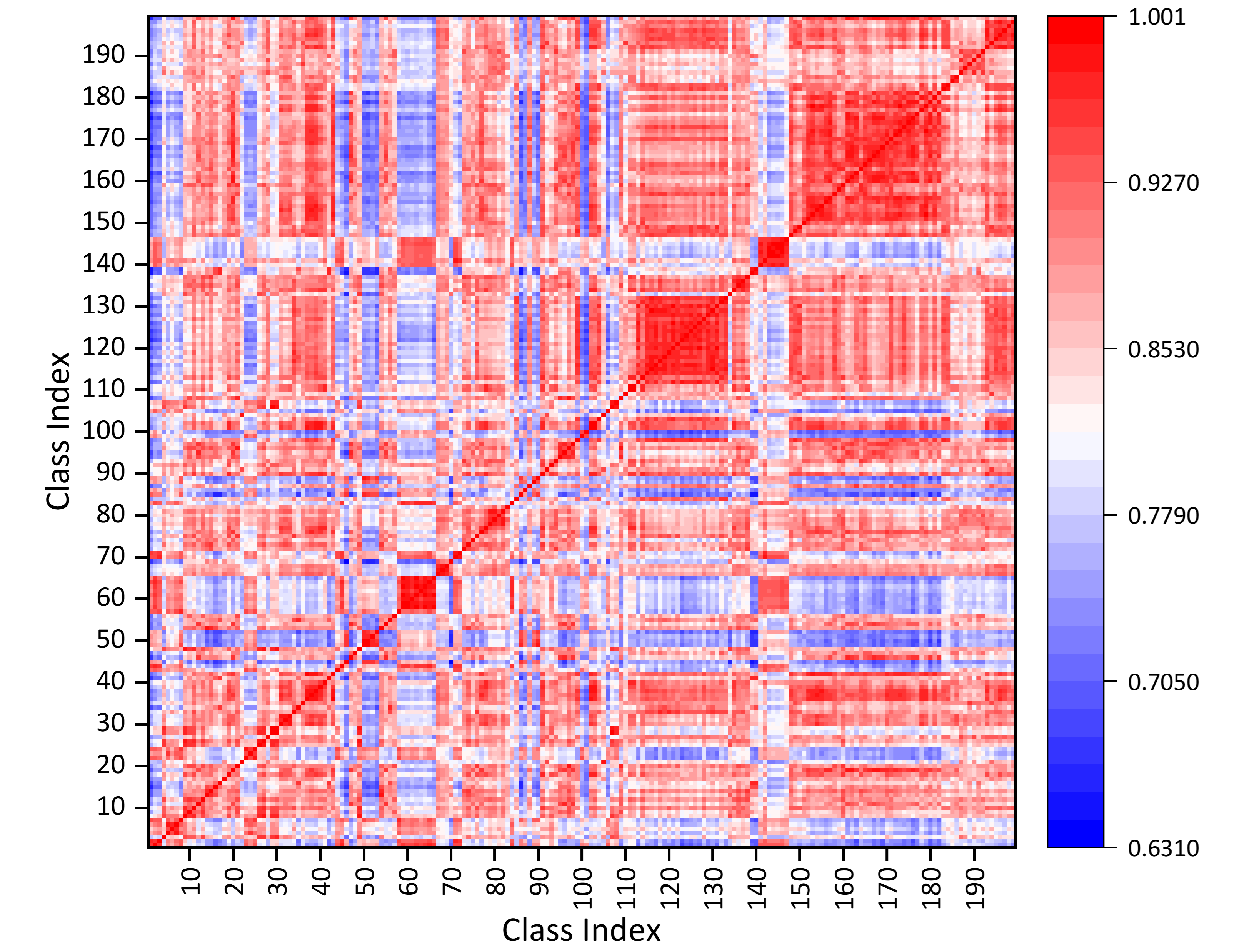}
    \caption{Cosine similarities between calculated prototypes.}
    \label{fig:proto_sim}
  \end{subfigure}
  \caption{The left figure validates that our extracted features fit better to the downstream task compared to features extracted by frozen CLIP. The right figure illustrates the correlations between prototypes. The red color indicates a high similarity, which often happen in close class indices. This is consistent with the the CUB-200-2011 dataset \cite{wah2011caltech}, where classes of close semantic meanings are indexed close to each other.}
  \label{fig:short}
\end{figure}

\textit{Importance of the prototype based feature space alignment.} The motivation for using prototype based feature space alignment instead of directly aligning the feature of each sample from the trainable feature extractor with the feature from the frozen CLIP's image encoder is: it offers the flexibility of obtaining features in the same feature space while being more class-discriminative regarding the specific downstream task to further boost the performance. We compare the performance of our prototype based feature space alignment described in equation \ref{eq:alignment} with the sample-wise alignment loss, which is calculated as:
\begin{equation}
 \mathcal{L}_{s\_align} =  
 \frac{ f_{clip}(\mathbf{x}_i)(g(\mathbf{x}_i))^T}{\|f_{clip}(\mathbf{x}_i)\|\|g(\mathbf{x}_i)\|}. 
\end{equation}
We set $\beta_1=\beta_2=1$ for this experiment.
The accuracy of using the above sample-wise alignment loss is 79.8\%, while our proposed prototype based alignment loss reaches 81.8\%, validating our design choice. An illustrative figure visualizing the high-dimensional features extracted by the frozen CLIP encoder and our trained feature extractor from 10 random classes using t-SNE \cite{van2008visualizing} (a tool visualizing high-dimensional features) is offered in Figure \ref{fig:t_sne}. This figure shows that our feature extractor successfully obtains features that are more compact, further validating the benefit of aligning towards a set of fixed class prototypes. Figure \ref{fig:proto_sim} offers the analysis on the similarity between calculated fixed prototypes, where red colors indicate high similarities. The ``block-wise'' similarity pattern observed in the figure is consistent with the dataset's semantics, as the CUB-200-2011 dataset \cite{wah2011caltech} has classes of similar semantics in neighboring class indices.

\textit{Influence of the alignment strength.}
We ablate the hyperparameter of $\beta_2$ by setting $\beta_2=1$ and $\beta_2=5$ while keeping the $\beta_1=1$ for a fair comparison. The performance when using $\beta_2=1$ is 81.8\% while the performance using $\beta_2=5$ reaches 83.3\%, indicating that a stronger feature space alignment enables the CLIP's frozen encoder to better guide the training of the trainable feature extractor. Note that simply setting $\beta_1=\beta_2=1$ readily yields a performance significantly better than the state-of-the-art Label-Free CBM \cite{oikarinen2023labelfree} by 7.5\%, suggesting the benefit of our overall framework design.

\textit{Limitations.}
How well the open vocabulary concepts match the corresponding visual features is limited by the quality of the feature space alignment of the pre-trained vision-language model, which is a common limitation in all prior CBM models incorporating CLIP \cite{radford2021learning}. However, we note that our proposed framework is applicable to any vision-language pre-trained model, as our design is agnostic to the pre-trained model's architecture. Thus we expect our model's capability to advance with the progress in vision-language pre-training.

\section{Conclusion}
In this work, we aim to design the first Concept Bottleneck Model with open vocabulary concepts to allow users to freely modify the concept set without re-training a new model. To this end, we propose a prototype based feature alignment strategy to align the feature space of a trainable feature extractor to that of the pre-trained vision-language model's visual space to indirectly align with the more human interpretable text space. We simultaneously train a classifier and propose to reconstruct the classification head via interpretable concepts to build the model in the form of CBM. Our ``OpenCBM'' not only offers a unique capability in flexible concept set modification, but also significantly outperforms the state-of-the-art CBM by 9\% in the classification accuracy.

\section*{Acknowledgments} 
This work was supported by the Hong Kong Innovation and Technology Fund (Project No. MHP/002/22), HKUST (Project No. FS111) and Project of Hetao Shenzhen-Hong Kong Science and Technology Innovation Cooperation Zone (HZQB-KCZYB-2020083).

\bibliographystyle{splncs04}
\bibliography{main}
\end{document}